\UseRawInputEncoding
\documentclass[letterpaper]{article}
\usepackage{flairs}%aaai
\usepackage{times}
\usepackage{helvet}
\usepackage{courier}
\usepackage[fleqn]{amsmath}
\usepackage{graphicx}
\usepackage{multirow}
\usepackage{appendix}
\begin{document}
% This file is an adoption of the style file for AAAI Press 
% proceedings, working notes, and technical reports.  This file is made 
% with minimal changes by explicit permission from AAAI.
\title{Localizing Adversarial Attacks To Produces More Imperceptible Noise}
\author{Pavan Reddy\\
The George Washington University\\
pavan.reddy@gwu.edu\\
\And Aditya Sanjay Gujral\\
The George Washington University\\
adityagujral@email.gwu.edu\\
}
\maketitle
\begin{abstract}
\begin{quote}
Adversarial attacks in machine learning traditionally focus on global perturbations to input data, yet the potential of localized adversarial noise remains underexplored. This study systematically evaluates localized adversarial attacks across widely-used methods, including FGSM, PGD, and C\&W, to quantify their effectiveness, imperceptibility, and computational efficiency. By introducing a binary mask to constrain noise to specific regions, localized attacks achieve significantly lower mean pixel perturbations, higher Peak Signal-to-Noise Ratios (PSNR), and improved Structural Similarity Index (SSIM) compared to global attacks. However, these benefits come at the cost of increased computational effort and a modest reduction in Attack Success Rate (ASR). Our results highlight that iterative methods, such as PGD and C\&W, are more robust to localization constraints than single-step methods like FGSM, maintaining higher ASR and imperceptibility metrics. This work provides a comprehensive analysis of localized adversarial attacks, offering practical insights for advancing attack strategies and designing robust defensive systems.
\end{quote}
\end{abstract}

\begin{figure}[h!]
    \centering
    \includegraphics[width=\columnwidth]{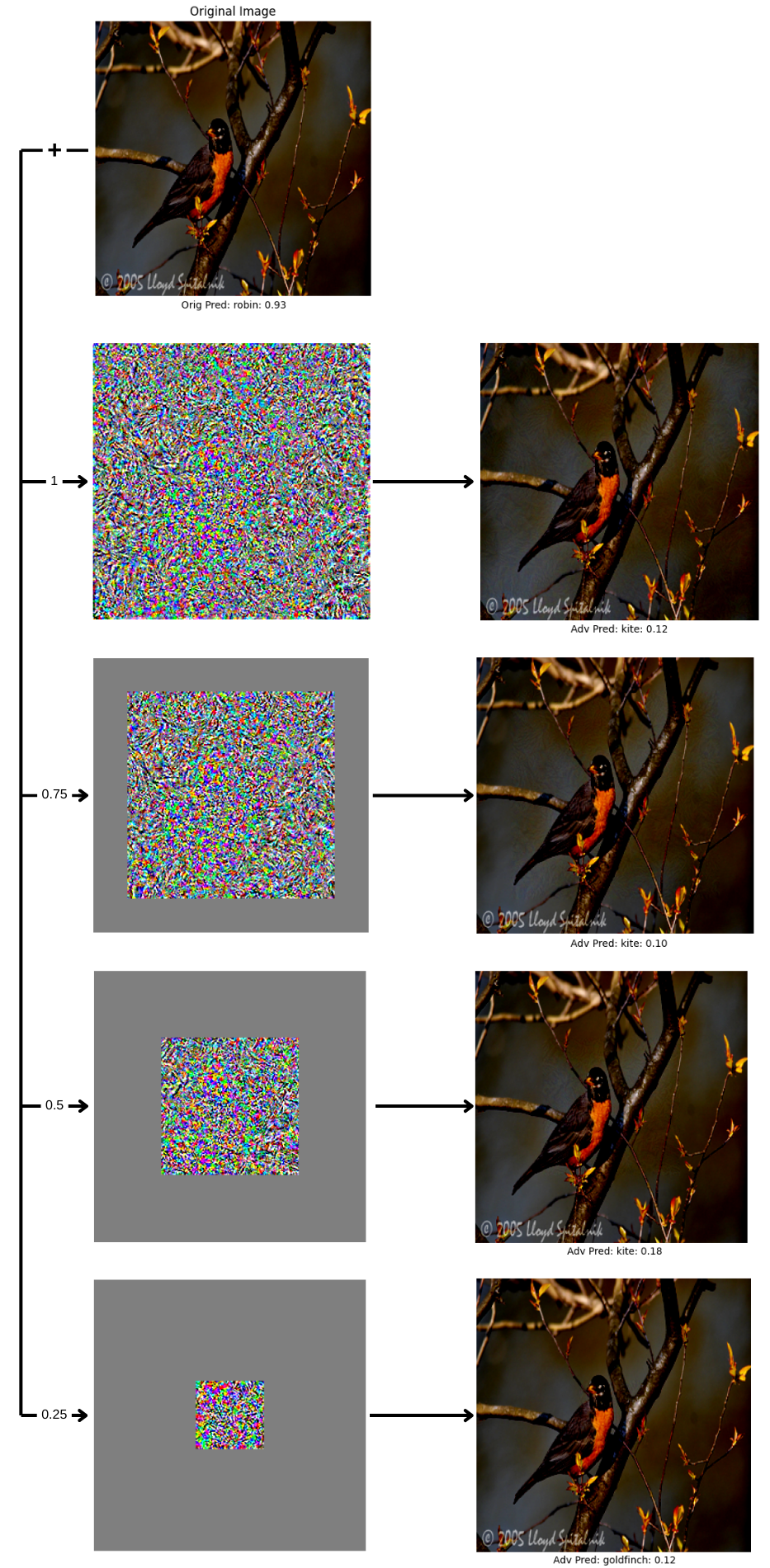} % Adjusts to column width
    \caption{Illustration of Projected Gradient Descent (PGD) Noise and Adversarial Prediction with localization parameter $(\gamma)$ 1, 0.75, 0.5, 0.25}
    \label{fig:attacks}
\end{figure}

\section{Introduction}
Deep Neural Networks are increasingly used in critical domains such as healthcare, finance, and autonomous systems \cite{ApplicationsDL}. This widespread adoption raises security concerns, as adversarial attacks \cite{goodfellow2015explainingharnessingadversarialexamples} small, crafted input perturbations—can cause models to fail unpredictably \cite{SecurityRisks}. Improving attack methods is essential for developing effective defenses \cite{Biggio}.

Most research focuses on global attacks that perturb entire images \cite{Chakraborty}, while localized attacks remain less explored \cite{localized} \cite{lavan}. Even methods like the Pixel Attack \cite{pixel}, though minimal in scope, are applied globally.

This work investigates the effectiveness of localized adversarial attacks. While not a novel direction, we show that localized noise can better exploit the allowed \(\epsilon\)-range while reducing mean perturbation values. We also evaluate the runtime required to produce successful localized attacks.

\section{Technical Background}
Deep Neural Networks consist of interconnected neurons that process inputs through multiple layers to generate predictions. Small input changes can propagate through the network and alter the output. Adversarial attacks exploit this by introducing small perturbations that cause misclassification \cite{intriguingproperties}.

The objective is to find a perturbation \(\delta\) such that \(x' = x + \delta\) leads to \(f(x) \neq f(x')\), where \(f(\cdot)\) is the model's prediction function. Perturbations are typically constrained by a norm \(\|\delta\|_p \leq \epsilon\), where \(\epsilon\) is the perturbation budget. This leads to the constrained optimization problem:
\begin{equation}
    \min_{\delta} \, \mathcal{L}(f(x + \delta), y) \quad \text{subject to} \quad \|\delta\|_p \leq \epsilon.
    \label{eq:adv_attack_general}
\end{equation} 

In white-box settings \cite{goodfellow2015explainingharnessingadversarialexamples}, the attacker has access to model parameters and can identify sensitive pixels to induce misclassification with minimal changes. \\

\textbf{Fast Gradient Sign Method (FGSM)} \cite{goodfellow2015explainingharnessingadversarialexamples} perturbs the input in the direction of the sign of the gradient of the loss:
\begin{equation}
x' = x + \epsilon \cdot \text{sign}(\nabla_x J(x, y)),
\label{eq:FSGM_Original}
\end{equation}
where \(\nabla_x J(x, y)\) is the gradient of the loss with respect to \(x\). \\ 

\textbf{Projected Gradient Descent (PGD)} \cite{PGD} extends FGSM by iteratively applying small updates and projecting the result onto the \(\epsilon\)-ball:
\begin{equation}
x_{t+1} \leftarrow \Pi_\epsilon \left( x_t + \alpha \cdot \text{sign} \left( \nabla_x J(x_t, y) \right) \right),
\label{eq:PGD_Original}
\end{equation}
where \(\Pi_\epsilon\) projects onto the \(L_\infty\)-ball centered at \(x\). \\

\textbf{Carlini \& Wagner (C\&W) Attack} \cite{CW} formulates an optimization to minimize perturbation while achieving misclassification:
\begin{equation}
x = \text{argmin}_{x} \left( C \cdot g(x) + \|\delta\|_2^2 \right),
\label{eq:CW_Original}
\end{equation}
with update step:
\begin{equation}
x_{t+1} = x_t - \eta \cdot \nabla_{x} \left( C \cdot g(x) + \|\delta\|_2^2 \right),
\label{eq:CW_Iteration}
\end{equation}
where \(g(x) = \max\left( \max_{i \neq y} Z(x)_i - Z(x)_y, -\kappa \right)\) controls the misclassification margin.

\section{Related Work}

While most research focuses on non-localized attacks, localized adversarial attacks have also been explored. These attacks perturb specific image regions and are often less perceptible.

\textit{Adversarial Patch} \cite{brown2018adversarialpatch} introduced a universal, localized patch that reliably causes misclassification in both digital and physical settings. 

\textit{LaVAN} \cite{lavan} showed that visible perturbations in small image regions, even away from the main object, can mislead DNNs.

\textit{Localized Uncertainty Attacks} \cite{dia2021localizeduncertaintyattacks} targeted uncertain regions in a model’s decision space, producing subtle yet effective perturbations.

\cite{localized} demonstrated that regional attacks reduce the \(L_p\) norm while preserving transferability, balancing efficiency and imperceptibility.

\section{Methodology}

\subsection{Method}

Our method follows \cite{localized}, assuming full access to model parameters and output probabilities \(p(y|x)\). We apply adversarial attacks by modifying the input as:
\begin{equation}
   x_l = x + (N \odot M),
   \label{eq:localized_attack}
\end{equation}
where \(N\) is a noise tensor and \(M\) is a binary mask defining the localized region. Gradients are restricted to the masked area, allowing targeted perturbations.

This method integrates into FGSM, PGD, and C\&W attacks by optimizing over \(N\):

\begin{align}
N &= N + \epsilon \cdot \text{sign}(\nabla_N J(x_l, y)) \label{eq:fgsm_loc} \\
N_{t+1} &\leftarrow \Pi_\epsilon \left( N_t + \alpha \cdot \text{sign} \left( \nabla_N J(x_l, y) \right) \right) \label{eq:PGD_loc} \\
N_{t+1} &= N_t - \eta \cdot \nabla_{x_l} \left( C \cdot g(x_l) + \|N\|_2^2 \right) \label{eq:CW_loc}
\end{align}

\subsection{Experimental Setup}

\begin{table*}[h!]
\centering
\begin{tabular}{|p{2cm}|p{1.5cm}|p{1.5cm}|p{1.5cm}|p{1.7cm}|p{1.5cm}|p{1.5cm}|p{1.5cm}|}
\hline
\textbf{Attack} & \textbf{Mask ($\gamma$)} & \textbf{ASR (\%)} & \textbf{Average Mean Change (\%)} & \textbf{Average PSNR Change (\% $\times 10^{-5}$)} & \textbf{Average SSIM Change (\%)} & \textbf{Average DR Change (\%)} & \textbf{Average Iterations Change (\%)} \\ \hline
\multirow{4}{*}{FSGM} & 1.00 & 0.76 & 0.00   & 0.00 & 0.00  & 0.00 & - \\ \cline{2-8}
                      & 0.75 & 0.72 & -43.37 & 0.57 & 4.00  & 0.00 & - \\ \cline{2-8}
                      & 0.50 & 0.60 & -74.47 & 1.06 & 7.00  & 0.00 & - \\ \cline{2-8}
                      & 0.25 & 0.22 & -93.71 & 1.50 & 8.66  & 0.00 & - \\ \hline
\multirow{4}{*}{PGD}  & 1.00 & 1.00 & 0.00   & 0.00 & 0.00  & 0.00 & 0.00 \\ \cline{2-8}
                      & 0.75 & 1.00 & -20.71 & -0.16 & -0.02 & 41.05  & 65.61 \\ \cline{2-8}
                      & 0.50 & 0.84 & -42.63 & -0.53 & -0.01 & 83.26  & 392.09 \\ \cline{2-8}
                      & 0.25 & 0.36 & -70.33 & 0.40 & 0.12  & 288.64 & 2257.69 \\ \hline
\multirow{4}{*}{CW}   & 1.00 & 0.82 & 0.00   & 0.00 & 0.00  & 0.00 & 0.00 \\ \cline{2-8}
                      & 0.75 & 0.88 & -16.22 & -0.01 & -0.00 & 11.88  & 6.33 \\ \cline{2-8}
                      & 0.50 & 0.82 & -41.88 & -0.01 & 0.00  & 25.03  & 33.42 \\ \cline{2-8}
                      & 0.25 & 0.72 & -60.43 & -0.03 & 0.01  & 114.09 & 589.27 \\ \hline
\end{tabular}
\caption{Evaluation of FGSM, PGD, and C\&W attacks under varying mask sizes ($\gamma = 1.0, 0.75, 0.5, 0.25$). Metrics include ASR, perturbation magnitude (Mean Change), PSNR, SSIM, Dynamic Range (DR), and average iterations required for success.}
\label{tab:results}
\end{table*}

We evaluated localized FGSM, PGD, and C\&W attacks on the InceptionV3 model pretrained on ImageNet using 50 images from 9 classes. All images were resized to \(299 \times 299 \times 3\) and preprocessed with TensorFlow's standard Inception function.

\noindent \textbf{Attack Success Criteria:}
\begin{itemize}
    \item \textit{FGSM:} Confidence of the predicted class reduced by at least 50\%.
    \item \textit{PGD/C\&W:} Misclassification within 250 iterations.
\end{itemize}

\noindent \textbf{Attack Parameters:}
\begin{itemize}
    \item \textit{FGSM:} \(\epsilon = 0.05\)
    \item \textit{PGD:} \(\epsilon = 0.02\), \(\alpha = 0.01\)
    \item \textit{C\&W:} Learning rate = 0.01, \(\kappa = 1000\), \(C = 10\)
\end{itemize}

\noindent \textbf{Masking Strategy:} 
We define \(\gamma\) as the fraction of active pixels, with tested values: \(1.0\), \(0.75\), \(0.5\), and \(0.25\), corresponding to decreasing central region sizes.

\subsection{Metrics}

\textbf{Attack Success Rate (ASR):} Fraction of successful attacks over total attempts. \\

\noindent \textbf{Number of Iterations:} Steps required to achieve misclassification (max 250). \\

\noindent \textbf{Mean Pixel Value(MPV):} Average absolute intensity of perturbation:
\[
\text{MPV} = \frac{\sum |N|}{H \times W \times C}
\]

\noindent \textbf{Peak Signal to Noise Ratio(PSNR):} Log-scaled ratio between maximum pixel intensity and noise energy:
\[
\text{PSNR} = 10 \cdot \log_{10}\left(\frac{\text{MAX}^2}{N}\right)
\]

\noindent \textbf{Dynamic Range (DR):} Range of values in the perturbation:
\[
\text{DR} = \max(N) - \min(N)
\]

\noindent \textbf{SSIM:} Structural similarity between original and perturbed images:
\[
\text{SSIM}(x, y) = \frac{(2 \mu_x \mu_y + C_1)(2 \sigma_{xy} + C_2)}{(\mu_x^2 + \mu_y^2 + C_1)(\sigma_x^2 + \sigma_y^2 + C_2)}
\]
Here, \(\mu_x\), \(\mu_y\) are the mean intensities, \(\sigma_x^2\), \(\sigma_y^2\) the variances, \(\sigma_{xy}\) the covariance, and \(C_1\), \(C_2\) small constants for numerical stability.

\section{Results}

\begin{figure*}[h!]
    \centering
    \includegraphics[width=0.85\textwidth]{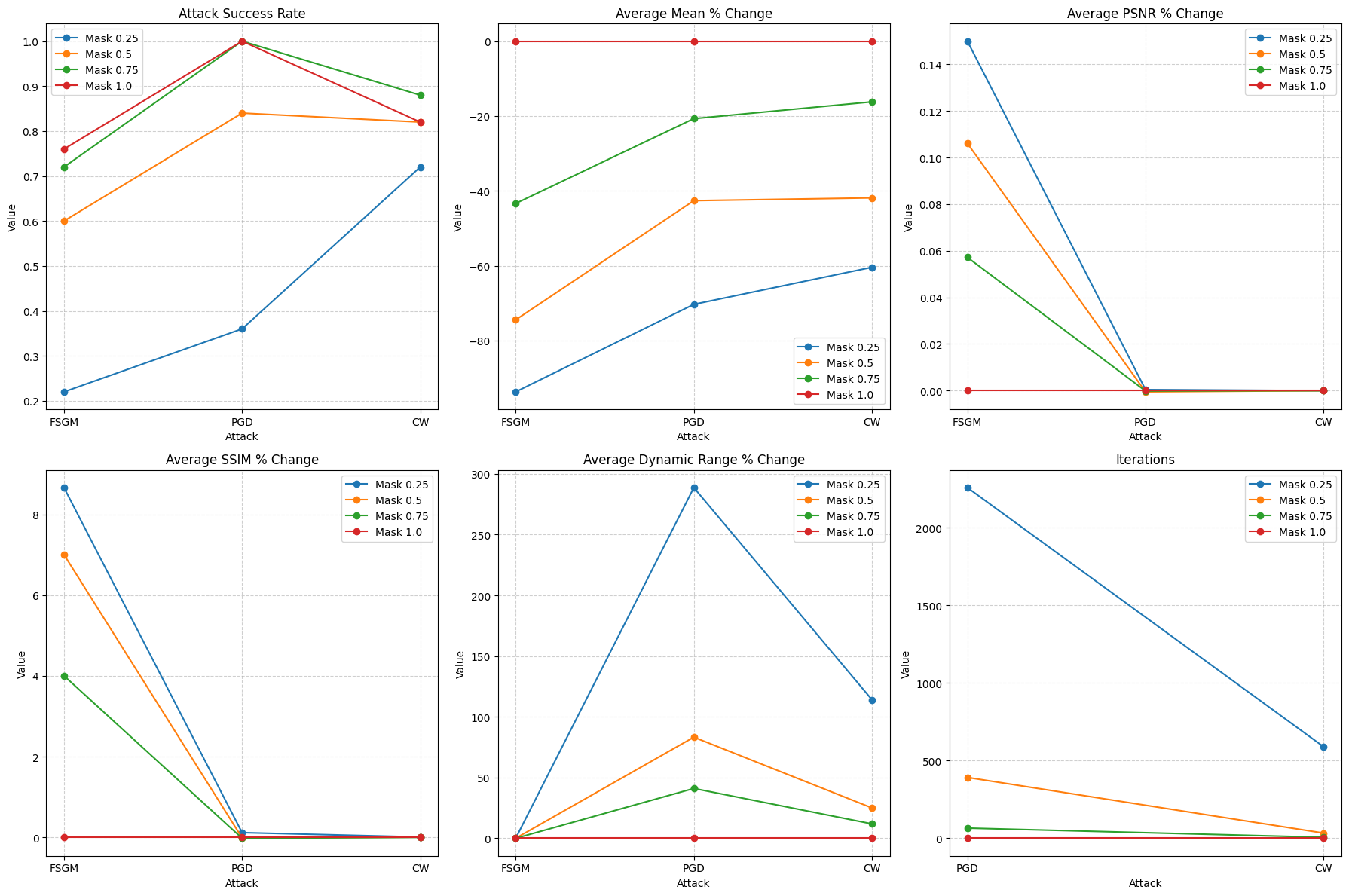}
    \caption{Evaluation of FGSM, PGD, and C\&W attacks across mask sizes (\(\gamma = 1.0, 0.75, 0.5, 0.25\)) using six metrics: ASR, Mean Pixel Value, PSNR, SSIM, DR, and Iterations.}
    \label{fig:graphs}
\end{figure*}

In this section, we evaluate the proposed mask-based method for localized adversarial attacks. Experiments were conducted under varying masking constraints (\(\gamma = 1.0, 0.75, 0.5, 0.25\)) to assess effectiveness, imperceptibility, and computational efficiency.

\subsection{Attack Success Rate (ASR)}

ASR measures the percentage of adversarial examples that achieve the intended misclassification or confidence reduction. As shown in Table~\ref{tab:results}, ASR decreases as \(\gamma\) is reduced:
\begin{itemize}
    \item \textbf{FGSM:} ASR drops from \(76\%\) at \(\gamma=1.0\) to \(22\%\) at \(\gamma=0.25\), reflecting reduced effectiveness with smaller perturbation regions.
    \item \textbf{PGD:} Achieves \(100\%\) ASR at \(\gamma=1.0\) and \(\gamma=0.75\), but falls to \(36\%\) at \(\gamma=0.25\).
    \item \textbf{C\&W:} Maintains relatively high ASR (\(72\%\)–\(88\%\)) across all mask sizes, indicating robustness to localization constraints.
\end{itemize}

\subsection{Perturbation Imperceptibility}

Imperceptibility was evaluated using Mean Pixel Value, PSNR, and SSIM. As \(\gamma\) decreases, fewer pixels are perturbed, resulting in subtler modifications:
\begin{itemize}
    \item \textbf{Mean Pixel Value:} Drops significantly with smaller masks—by \(93.71\%\) in FGSM, \(70.33\%\) in PGD, and \(60.43\%\) in C\&W at \(\gamma=0.25\).
    \item \textbf{PSNR:} Increases for FGSM due to bounded noise, but decreases for PGD and C\&W, indicating sharper perturbation peaks at smaller \(\gamma\).
    \item \textbf{SSIM:} Higher SSIM is observed for FGSM at low \(\gamma\); PGD shows inconsistent trends; C\&W maintains stable similarity across mask sizes.
\end{itemize}

\subsection{Dynamic Range of Perturbations}

As \(\gamma\) decreases, the dynamic range of the noise increases:
\begin{itemize}
    \item \textbf{PGD:} \(+288.64\%\) increase in range at \(\gamma=0.25\) compared to \(\gamma=1.0\).
    \item \textbf{C\&W:} \(+114.09\%\) increase under the same condition.
\end{itemize}
This suggests that more extreme values are used to compensate for the limited spatial region available.

\subsection{Computational Efficiency}

As \(\gamma\) is reduced, more iterations are required to achieve successful attacks:
\begin{itemize}
    \item \textbf{PGD:} Requires \(2257.69\%\) more iterations at \(\gamma=0.25\) than at \(\gamma=1.0\).
    \item \textbf{C\&W:} Iterations increase by \(589.27\%\) at \(\gamma=0.25\), indicating higher computational cost for localized optimization.
\end{itemize}

\subsection{Attack Comparisons}

\begin{itemize}
    \item \textbf{FGSM:} Suffers from low ASR and high perceptibility with smaller masks.
    \item \textbf{PGD:} Highly effective at larger \(\gamma\), but becomes less efficient and less successful as the mask size shrinks.
    \item \textbf{C\&W:} Achieves the best balance, maintaining high ASR and good imperceptibility across all masking levels.
\end{itemize}

Figure~\ref{fig:attacks} illustrates PGD adversarial examples. As \(\gamma\) decreases, perturbations become more concentrated and less visible, consistent with metric trends.

\section{Discussion}

This study shows that localized adversarial noise improves imperceptibility while preserving attack effectiveness, especially in iterative methods like PGD and C\&W. Localized noise yields lower mean pixel values, higher PSNR, and better SSIM than global noise. However, smaller masks (\(\gamma\)) reduce ASR and increase computation—e.g., PGD needed 2257\% more iterations at \(\gamma = 0.25\) than at \(\gamma = 1.0\).

FGSM performed poorly under localization, with ASR dropping from 76\% to 22\%, highlighting its limitations. PGD and C\&W were more robust, with C\&W maintaining stable ASR and visual quality across all \(\gamma\).

Our results are based on a single model (InceptionV3) and dataset (ImageNet), limiting generalizability. Broader evaluation across architectures like ResNet, VGG, and ViT is needed. Also, the impact of localized noise on existing defense methods remains unexplored and should be evaluated.

\subsection{Future Work}

Future work includes adaptive masking to better balance stealth, success, and efficiency \cite{tramer2020adaptiveattacksadversarialexample}. Evaluating localized attacks on text, audio, and tabular data can expand applicability. Theoretical study of optimization behavior—e.g., gradient flow and decision boundaries—may clarify why small \(\gamma\) reduces ASR. Localized noise may also support targeted manipulation and help evade detection \cite{DetectingLocalized}.

\section{Conclusion}

We evaluated localized adversarial noise in FGSM, PGD, and C\&W attacks. Localized perturbations improve imperceptibility as measured by Mean Pixel Value, but reduce ASR and increase cost at smaller mask sizes.

FGSM degrades under localization, while PGD and especially C\&W retain higher ASR and visual quality. These findings support iterative methods for localized attacks where stealth is critical.

Localized noise has uses in robustness testing, security evaluation, and detection evasion. Future work should explore dynamic masking, defense comparisons, cross-model validation, and deeper theoretical analysis to optimize trade-offs.

\bibliographystyle{flairs}
\bibliography{references}

\end{document}